\documentclass[11pt]{article}

\usepackage[final]{acl}

\usepackage{times}
\usepackage{latexsym}
\usepackage[T1]{fontenc}
\usepackage[utf8]{inputenc}
\usepackage{microtype}
\usepackage{inconsolata}
\usepackage{graphicx}

\usepackage[most]{tcolorbox}
\usepackage{longtable}

\usepackage{placeins}
\usepackage{listings}
\usepackage{appendix}
\usepackage{enumitem}
\usepackage{tikz}
\usepackage{soul}
\usepackage{subcaption}
\usepackage{xspace}

\usetikzlibrary{shapes.geometric, arrows, fit}
\usetikzlibrary{decorations.pathmorphing} 
\usetikzlibrary{decorations.pathreplacing} 

\setitemize{noitemsep,topsep=5pt,parsep=2pt,partopsep=0pt}

\newcommand\acronym{LLM-RBMT\xspace}

\newcommand{\displayTranslation}[9]{%
    \begin{tcolorbox}[ %
        title={#1}, %
        colback=white, %
        colframe=black, %
        colbacktitle=lightgray, %
        coltitle=black, %
        fonttitle=\bfseries, %
        enhanced, %
        attach boxed title to top left={yshift=-2mm, xshift=2mm}, %
        ] %
        {\small \textbf{Simple} \hfill {\color{darkblue} \textbf{Semantic Similarity}}}\\
        {#2} \hfill {\color{darkblue} #3} \\
        {\small \textbf{Comparator}} \\
        {#4} \hfill {\color{darkblue} #5} \\
        {\small \textbf{Target}} \\
        {#6} \\
        {\small \textbf{Backwards}} \\
        {#7} \hfill {\color{darkblue} #8}%
        \begin{center}%
            {\small (model: #9)}%
        \end{center}%
    \end{tcolorbox}%
}%

\title{LLM-Assisted Rule Based Machine Translation for Low/No-Resource Languages}

\author{Jared Coleman \\ {\bf Bhaskar Krishnamachari} \\ {\bf Khalil Iskarous} \\
  University of Southern California\\
  \texttt{{\{jaredcol,bkrishna,kiskarou\}@usc.edu}} \\\And
  Ruben Rosales \\
  \texttt{rubenrosales@ieee.org}
}

\date{\today}

\begin{document}
\maketitle

\begin{abstract}
We propose a new paradigm for machine translation that is particularly useful for no-resource languages (those without any publicly available bilingual or monolingual corpora): \acronym (LLM-Assisted Rule Based Machine Translation). Using the \acronym paradigm, we design the first language education/revitalization-oriented machine translator for Owens Valley Paiute (OVP), a critically endangered Indigenous American language for which there is virtually no publicly available data. We present a detailed evaluation of the translator's components: a rule-based sentence builder, an OVP to English translator, and an English to OVP translator. We also discuss the potential of the paradigm, its limitations, and the many avenues for future research that it opens up.
\end{abstract}

\section{Introduction}\label{sec:intro}
Large Language Models like OpenAI's GPT series~\cite{gpt:gpt4-tech-report} have shown remarkable capability at an impressively wide range of natural language tasks~\cite{gpt:agi} including machine translation~\cite{trans:eval}.
These models work because they are trained on vast amounts of natural language data, primarily from the internet~\cite{gpt:gpt4-tech-report}.
For languages that are low-resource (languages for which there is little publicly available data) or no-resource (languages for which there is \textit{no} publicly available data), models like these do not perform well on their own~\cite{palm,gpt:low-resource-translation}.
There have been many efforts in improving machine translation for low-resource languages (see~\cite{low_resource_survey} for a comprehensive survey), but no-resource languages have received much less attention in the literature.

Over past decades, researchers and community leaders have led many remarkable efforts in endangered language revitalization~\cite{CoronelMolina2016IndigenousLR,taylor2022access} and reclamation~\cite{baird}.
In this paper, we begin to explore how the impressive general-purpose language skills of LLMs might be helpful for these kinds of efforts by introducing a new paradigm for low/no-resource machine translation: \acronym (LLM-Assisted Rule-Based Machine Translation).
The intuition behind our approach is that, given the correct tools, humans are good at translating simple sentences even for languages they do not know.
For example, a common task for students in a language class is to translate sentences given a set of words and rules.
Given a conjugation table and a set of verbs 
a motivated student could probably translate a list of simple subject-verb (e.g., ``I eat'', ``you sing'', etc.) sentences with extremely high accuracy.
Of course, they are limited to only translating specific kinds of sentences with a limited vocabulary.
Still, the idea is interesting: if we provide enough context to an LLM like ChatGPT, which has been shown to exhibit human-level performance on many natural language tasks, we may not need it to actually \textit{know} (exhibit general fluency in) the target language we want to translate to/from.
While rule-based translation can likely never achieve the quality of modern ML translators for high-resource languages, this technique has a few potential advantages for low/no-resource languages.
First and most importantly, it requires no parallel corpus.
Also, it is amenable to \textit{partial translation}.
Humans (and, as we'll show in this paper, LLMs) are capable of giving partial translations when they don't have all of the necessary vocabulary.
For example, a Spanish student that doesn't know the word for ``dog'' might still be able to translate the rest of the sentence ``the dog ate the apple yesterday'' by saying ``el [dog] comió la manzana ayer''.

In this paper, we leverage LLMs (specifically, gpt-3.5-turbo and gpt-4 from OpenAI's GPT series) to break down natural language sentences into structured simple sentences compatible with hand-crafted rule-based translators.
We also use them to turn sentence structured information (in JSON format) into natural language sentences.
Using this approach, the LLM never interacts directly with the target language.
Rather, we rely on the LLM to tell us how to use the simple, rule-based translators to provide a translation as close as possible to the user's original input.
We use this technique to build and evaluate two machine translation tools for Owens Valley Paiute (OVP, also called Eastern Mono or Monache in linguistics literature, \texttt{ISO 639-3 mnr}~\cite{mnr}), a critically endangered Indigenous American language in the Uto-Aztecan language family~\cite{atlas_endangered_languages}.
The first is a selection-based OVP to English simple sentence translator and the second is an English to OVP translator that uses available vocabulary to construct translations of arbitrary user input sentences.
The translators are oriented toward language teaching and revitalization.
They are not designed to be general-purpose translators, but rather as tools to help no-resource language learners express ideas using simple sentence constructions.

\subsection{Contributions}
The main contributions of this work are\footnote{The code behind all contributions is open-source: \url{https://github.com/kubishi/kubishi_sentences}}:
\begin{enumerate}
    \item An extensible OVP sentence builder for constructing valid subject-verb and subject-verb-object sentences in Owens Valley Paiute.
    \item An LLM-assisted OVP to English translator that translates OVP sentence builder sentences to English with high accuracy.
    \item An LLM-assisted English to OVP translator that translates English sentences to Owens Valley Paiute using the sentence-builder and OVP to English translation tools.
    \item A novel methodology for the design and evaluation of no-resource language translators.
\end{enumerate}
The rest of this paper is organized as follows.
We discuss related work in Section~\ref{sec:related}.
We present the sentence building tool and OVP to English translation system in Section~\ref{sec:ovp2eng}.
Then, in Section~\ref{sec:eng2ovp}, we present the English to OVP translation system and report results on translation quality for different types of input sentences using embeddings models to measure semantic similarity.
We conclude the paper with a summary of contributions and discussion of future research directions in Section~\ref{sec:conclusion}.

\section{Related Work}\label{sec:related}
The landscape of low-resource machine translation is vast and constantly growing.
A comprehensive survey on this subject is provided by \cite{low_resource_survey}, which outlines the current techniques, guidance for selecting which techniques to use for a given language, and future directions for research. 
Of particular interest within this survey is the examination of unsupervised neural machine translation.
While there exists promising research on constructing translators from minimal corpora, these methods invariably require \textit{some} natural language data and thus have limited applicability to no-resource languages (such as OVP). 
The survey also discusses data augmentation strategies, including word or phrase replacement-based augmentation and Back-Translation-based Data Augmentation, both of which could potentially be integrated with some of the solutions presented in this paper (in particular the sentence builder and OVP to English translator to be presented in Section~\ref{sec:ovp2eng}).
Such an exploration is an interesting topic for future work.
Other approaches discussed in the survey such as supervised learning, transfer learning, and semi-supervised learning are inapplicable to our scenario due to the absence of bilingual or monolingual corpora.

Contrary to the prevailing assumption in the literature that rule-based machine translation (RBMT) is a relic of the past, there remains active research and development in RBMT systems tailored for the most under resourced of  languages~\cite{rbmt:aperterium, rbmt:workflows, rbmt:low_resource}.
Recent work has also explored the utilization of Large Language Models (LLMs) for enhancing translation capabilities in low-resource languages through fine-tuning techniques~\cite{low_resource_finetune_irish}.
Although this approach has shown promise in improving LLM translation quality for low-resource languages like (e.g., Irish), its reliance on bilingual corpora make it infeasible for no-resource languages like OVP.
Recently, semantic similarity has been used to evaluate the quality of Machine Translation systems~\cite{ss:testing_mt,ss:eval_mt}.
We observe that this technique is particularly useful for evaluating the quality of the English to OVP translator presented in this paper, due to the lack of parallel corpora.
See~\cite{ss:mteb} for a comprehensive benchmark of different embeddings models used for computing semantic similarities.

\section{OVP to English Translation}\label{sec:ovp2eng}
In this section, we present an LLM-assisted selection-based OVP to English translator.
The first piece of this translation system is a selection-based OVP sentence builder.
The sentence builder allows users to select each of the following parts of speech from a list of choices until they form a valid sentence:
\begin{itemize}
    \item Subject: The subject of the sentence.
    \item Subject Suffix: In OVP, noun subjects are always suffixed with either \textbf{-ii} (if the subject is proximal to the speaker) or \textbf{-uu} (if the subject is distant to the speaker).
    \item Verb: the verb of the sentence.
    \item Verb Suffix: In OVP, verb suffixes express the tense/aspect of the verb. 
    \item Object: The object of the sentence (disallowed if the selected verb is intransitive, optional otherwise).
    \item Object Suffix: In OVP, noun objects are always suffixed with either \textbf{-(n)eika} (if the subject is proximal to the speaker) or \textbf{-(n)oka} (if the subject is distant to the speaker).
    \item Verb Object Pronoun Prefix: In OVP, object pronouns are prefixed to the verbs they are the object of. Object pronouns are required for all transitive verbs. Even when a noun object is specified, the pronoun is still required and should match the object suffix (\textbf{-(n)eika} matches \textbf{a-} or \textbf{ma-} and \textbf{-(n)oka} matches \textbf{u-}).
\end{itemize}
Not all of these parts of speech are required to create a valid sentence in OVP, though.
In fact, some of them are incompatible.
For example, if an intransitive verb is chosen, then it cannot have an object or object pronoun.
In other words, the valid choices a user can make is a function of the choices they have already made.
In our python implementation for this translator, we process each user selection and change the list of valid options for each part-of-speech based on their current selections to ensure they always create a valid OVP sentence.
This is the \textit{rule} part of LLM-Assisted \textit{Rule}-Based Machine Translation that requires expert knowledge in the target language to implement.
The entire vocabulary available for the users to select among fits on a single page (and can be found in Appendix~\ref{apx:vocab}).

After the user creates a valid OVP sentence, we translate it by first encoding the following sentence information into an English-only (using vocabulary definitions) \textit{structured} simple sentence:
\begin{itemize}
    \item Subject: noun or pronoun subject in English
    \item Subject Proximity: \textbf{proximal} if user selected subject suffix \textbf{-ii} or \textbf{distant} if the user selected subject suffix \textbf{-uu}
    \item Object: noun or pronoun object in English
    \item Object Proximity: \textbf{proximal} if user selected object suffix \textbf{-(n)eika} or \textbf{distant} if the user selected object suffix \textbf{-(n)oka}
    \item Verb: verb in English
    \item Verb Tense/Aspect: one of \textbf{past}, \textbf{present}, \textbf{future}, \textbf{past-continuous}, \textbf{present-continuous}, \textbf{present-perfect}
\end{itemize}
Then, we use few-shot prompting to encourage an LLM to transform the structured English data into a natural language sentence.
Consider the example in Figure~\ref{fig:few-shot:segment} where few-shot training examples (colored black) tell the LLM how to respond to the actual structured data for the randomly generated sentence for ``Wo'ada-ii pagwi-noka u-zawa-dü.'' (colored blue).
Observe, the LLM is prompted to translate using only the English, structured version of the selected sentence.
\begin{figure*}
\begin{tcolorbox}[
    title=Wo'ada-ii pagwi-noka u-zawa-dü.,
    colback=white,
    colframe=black,
    colbacktitle=lightgray,
    coltitle=black,
    fonttitle=\bfseries,
    enhanced,
    attach boxed title to top left={yshift=-2mm, xshift=2mm},
    ]
    {\small \color{black!30!white} system}
    
    You are an assistant for translating structured sentences into simple natural English sentences.
    
    {\small \color{black!30!white} user}
    
    [\{`part\_of\_speech': `subject', `positional': `proximal', `word': `wood'\}, \{`part\_of\_speech': `object', `positional': `proximal', `word': 'dog'\}, \{`part\_of\_speech': `verb', `tense': `present continuous (-ing)', `word': `see'\}]

    \hfill {\small \color{black!30!white} assistant}
    
    \hfill This wood is seeing this dog.
    
    
    
    
    
    {\small \color{black!30!white} user}

    [\{`part\_of\_speech': `subject', `positional': `distal', `word': `pinenuts'\}, \{`part\_of\_speech': `object', `positional': `distal', `word': `horse'\}, \{`part\_of\_speech': `verb', `tense': `future (will)', `word': `see'\}]
    
    \hfill \textcolor{black!30!white}{{\small assistant}} 
    
    \hfill Those pinenuts will see that horse.

    {\color{blue}
    {\small \color{blue} user}
    
    [\{`part\_of\_speech': `subject', `word': `mosquito', `positional': `proximal'\}, \{`part\_of\_speech': `object', `word': `fish', `positional': `distal'\}, \{`part\_of\_speech': `verb', `word': `cook', `tense': `present'\}]

    \hfill \textcolor{blue!30!white}{{\small assistant}} 
    
    \hfill This mosquito is cooking that fish.
    }
\end{tcolorbox}%
\caption{Few-shot examples for translating ``Wo'ada-ii pagwi-noka u-zawa-dü.'' using gpt-3.5-turbo.}
\label{fig:few-shot:segment}
\end{figure*}

To evaluate the accuracy of the translator, we generated 100 random valid OVP sentences by iteratively selecting a random choice among available choices for each of the parts of speech until the sentence is valid.
Of the 100 random sentences generated, 98 were translated into English accurately using gpt-3.5-turbo model from OpenAI's GPT-series.
Translations and their accuracy labels can be found in Appendix~\ref{apx:ovp2eng}.
While impressively accurate, this translator has many disadvantages.
It only works for simple subject-verb and subject-verb-object sentences that use the nouns and verbs available in the system.
Also, since pronouns and suffixes in OVP encode temporal/spatial information, translations don't always capture full meaning of the sentence.
The English translations are correct, but may be missing useful information.
For example "kamü-uu wo'abi-neika a-düka-ti" translates to "the jackrabbit is eating the worm", which is technically correct, but -uu also indicates that the jackrabbit is not present and the -neika indicates the worm \textit{is} present.
Then, since -ti can is used for both the present \textit{and} past continuous tenses (\textit{is [x]-ing} or \textit{was [x]-ing}), a better translation would be ``the jackrabbit was eating this worm''.
More advanced rules and better prompt-engineering may help mitigate this issue.

Despite some expected disadvantages, this translator has many advantages.
First, it is the first machine translator for OVP. It is also easy to extend the tool with new nouns and verbs.
Also, while implementing the rules required expert knowledge of what makes an OVP sentence valid, no expert knowledge of how the rules map to English  was required (or needed to be programmed), thanks to the LLM.
Finally, we believe this kind of translation system might be a useful educational tool that helps students learn how to build simple sentences.
It also may be useful as a data augmentation technique for training neural machine translation models for low-resource languages.

\section{English to OVP Translation}\label{sec:eng2ovp}
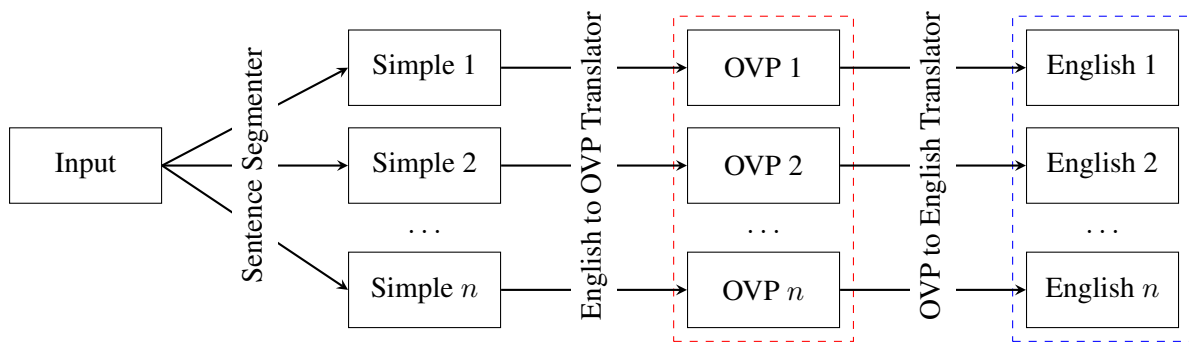
\begin{figure*}[!h]
    \centering
    
    \begin{tikzpicture}[node distance=2cm]
    \tikzstyle{box} = [rectangle, minimum width=2cm, minimum height=1cm, text centered, draw=black]
    \tikzstyle{arrow} = [thick,->,>=stealth]
    
    \node (origin) [box, draw=none] {};
    \node (simple1) [box, right of=origin, xshift=70] {Simple 1};
    \node (simple2) [box, below of=simple1, yshift=20] {Simple 2};
    \node (ldots1) [below of=simple2, yshift=32] {$\ldots$};
    \node (simple3) [box, below of=simple2, yshift=10] {Simple $n$};
    
    \node (input) [box, left of=simple2, xshift=-70, align=left] {Input};
    
    \node (ovp1) [box, right of=simple1, xshift=70] {OVP 1};
    \node (ovp2) [box, right of=simple2, xshift=70] {OVP 2};
    \node (ldots2) [right of=ldots1, xshift=70] {$\ldots$};
    \node (ovp3) [box, right of=simple3, xshift=70] {OVP $n$};
    
    \node (ovpbox) [draw=red, dashed, fit=(ovp1) (ovp2) (ovp3), inner sep=5pt] {};
    
    \node (en1) [box, right of=ovp1, xshift=70] {English 1};
    \node (en2) [box, right of=ovp2, xshift=70] {English 2};
    \node (ldots3) [right of=ldots2, xshift=70] {$\ldots$};
    \node (en3) [box, right of=ovp3, xshift=70] {English $n$};

    \node (ovpbox) [draw=blue, dashed, fit=(en1) (en2) (en3), inner sep=5pt] {};
    
    \draw [arrow] (input.east) -- (simple1.west);
    \draw [arrow] (input.east) -- (simple3.west);
    \draw [arrow] (input.east) -- (simple2.west) node[midway, fill=white, rotate=90] {Sentence Segmenter};
    
    \draw [arrow] (simple1.east) -- (ovp1.west);
    \draw [arrow] (simple3.east) -- (ovp3.west);
    \draw [arrow] (simple2.east) -- (ovp2.west) node[midway, fill=white, rotate=90] {English to OVP Translator};
    
    \draw [arrow, ] (ovp1.east) -- (en1.west);
    \draw [arrow, ] (ovp3.east) -- (en3.west);
    \draw [arrow, ] (ovp2.east) -- (en2.west) node[midway, fill=white, rotate=90] {OVP to English Translator};
    \end{tikzpicture}
    
    \caption{The entire English to OVP translation process. The box with a red, dashed border indicates the set of sentences in Owens Valley Paiute (the target language) and the box with a blue, dashed border indicates the set of English sentences they translate to. Ideally, the input sentence, simple sentences, and English output sentences will have equivalent or very similar semantic meaning.}
    \label{fig:architecture}
\end{figure*}

In this section, we present the first English to Owens Valley Paiute machine translator.
Unlike the OVP to English translator presented in the previous section, the user can input any sentence in natural language.
The translator works by first using an LLM to break the input sentence into a set of simple structured subject-verb and subject-verb-object sentences, discarding any adjectives, adverbs, prepositions, objects (except for direct objects), etc.
We encourage (through few-shot prompt engineering) the LLM to preserve as much semantic meaning as possible between the original sentence and the set of simple sentences.
Consider the example in Figure~\ref{fig:few-shot:eng2ovp} where few-shot examples (colored black) tell the LLM how to respond to a given input sentence ``We are playing and laughing.'' (colored blue)\footnote{We also leverage OpenAI's function calling capability to ensure that responses are consistently formatted. We refer interested readers to the open-source implementation's documentation at \url{https://github.com/kubishi/kubishi_sentences} for full details.}.
\begin{figure*}
\begin{tcolorbox}[
    title=We are playing and laughing.,
    colback=white,
    colframe=black,
    colbacktitle=lightgray,
    coltitle=black,
    fonttitle=\bfseries,
    enhanced,
    attach boxed title to top left={yshift=-2mm, xshift=2mm},
    ]
    {\small \color{black!30!white} system}
    
    You are an assistant that splits user input sentences into a set of simple SVO or SV sentences. The set of simple sentences should be as semantically equivalent as possible to the user input sentence. No adjectives, adverbs, prepositions, or conjunctions should be added to the simple sentences. Indirect objects and objects of prepositions should not be included in the simple sentences.
    
    {\small \color{black!30!white} user}
    
    I am sitting in a chair.

    \hfill {\small \color{black!30!white} assistant}
    
    \hfill [\{`subject': `I', `verb': `sit', `verb\_tense': `present\_continuous', `object': null\}]
    
    
    
    
    
    {\small \color{black!30!white} user}

    I saw two men walking their dogs yesterday at Starbucks while drinking a cup of coffee
    
    \hfill \textcolor{black!30!white}{{\small assistant}} 
    
    \hfill [\{`subject': `I', `verb': `see', `verb\_tense': `past', `object': `man'\}, \{`subject': `man', `verb': `walk', `verb\_tense': `past\_continuous', `object': 'dog'\}, \{`subject': `man', `verb': `drink', `verb\_tense': `past\_continuous', `object': `coffee'\}]

    {\color{blue}
    {\small \color{blue!30!white} user}
    
    We are playing and laughing.

    \hfill \textcolor{blue!30!white}{{\small assistant}} 
    
    \hfill [\{`subject': `we', `verb': `play', `verb\_tense': `present\_continuous', `object': null\}, \{`subject': `we', `verb': `laugh', `verb\_tense': `present\_continuous', `object': null\}]
    }
\end{tcolorbox}%
\caption{Few-shot training examples for the English to OVP using gpt-3.5-turbo.}
\label{fig:few-shot:eng2ovp}
\end{figure*}
Then, we use these structured sentences and available vocabulary to build valid OVP sentences with the sentence-building tool described in Section~\ref{sec:ovp2eng}.
Once the sentence is built, we use the translator described in Section~\ref{sec:ovp2eng} to translate the OVP sentences back into English.
The idea is that, while some meaning may have been lost between the original input sentence and the final output English sentences, the user can be fairly confident (given the accuracy of the OVP to English translator) that the final translations are correct.
The entire English to OVP translation process is depicted in Figure~\ref{fig:architecture}.

We evaluate the system by translating a set of 125 sentences. There are five types of sentences in the dataset (25 of each):
\begin{itemize}
    \item subject-verb (e.g., ``I read'' or ``she sings'')
    \item subject-verb-object (e.g., ``Mom made dinner'' or ``John read a book'')
    \item two-verb (e.g., ``She sings and dances.'' or ``I ate while watching TV.'')
    \item two-clause (e.g., ``My brother drove and I waited.'' or ``Harry wrote and Ron read.'')
    \item complex (e.g., ``Rachel and Monica share an apartment.'' or ``Romeo and Juliet loved deeply.'')
\end{itemize}
We translated all 125 sentences using both the gpt-3.5-turbo and gpt-4 models from OpenAI's GPT-series, resulting in a total of 250 translations.

To measure the quality of each translation, we compute the semantic similarity between the input sentence and:
\begin{itemize}
    \item the set of simple sentences generated by the LLM-powered segmenter (denoted \textit{simple}).
    \item the set of simple sentences with unknown vocabulary removed (denoted \textit{comparator}). Intuitively, this represents the best the translator can achieve given the vocabulary it has access to. For example, suppose the verb ``wash'' is not in our available vocabulary. Then the comparator sentence for the simple sentence ``The woman is washing'' would be ``The woman is [VERB]-ing'' 
    \item the ``round-trip'' English translation (denoted \textit{backwards}). This is the sentence produced using by translating the translated sentence (in OVP) to English using the method described in Section~\ref{sec:ovp2eng}.
\end{itemize}

The usefulness of these measurements depends greatly on the function used to compute semantic similarity.
We compute semantic similarity by generating embeddings for sentences (using some embeddings model) and computing the normalized cosine similarity between these embeddings.
For our application, we want semantically similar sentences to have small normalized cosine similarity, independent of other linguistic features like syntax.
For example, an ideal semantic similarity function would rank ``an apple is eaten by a man'' more similar to ``a man eats apples'' than the sentence ``a woman drinks coffee'', despite the latter sentence being essentially grammatically equivalent to the target sentence.

We evaluated seven different embeddings models for this purpose and measured the semantic similarity between twelve target sentences and a ranked list of 10 sentences for each ranging from most to least semantically similar (sentences can be found in Appendix~\ref{apx:semantic_similarity}).
For each target sentence, we compare the ground-truth ranking of the 10 sentences to the ranking determined by the semantic similarity scores yielded by a particular embeddings model.
We measure the similarity between the two rankings using two metrics: average displacement (average distance between a sentence's position in the computed ranking and its position in the target ranking) and RBO (Rank-biased Overlap~\cite{rbo}).
Table~\ref{tab:semantic} tabulates the results of this evaluation.
\begin{table*}[!ht]
\centering

\begin{tabular}{|lr|cc|cc|}
\hline
& & \multicolumn{2}{c|}{\textbf{Average}} & \multicolumn{2}{c|}{\textbf{}} \\
& & \multicolumn{2}{c|}{\textbf{Displacement}} & \multicolumn{2}{c|}{\textbf{RBO}} \\
\textbf{ Embeddings Model} & & \textbf{mean} & \textbf{std} & \textbf{mean} & \textbf{std} \\
\hline
\textbf{text-embedding-ada-002} & \cite{embeddings:text-embedding-ada-002} & 0.967 & 0.442 & 0.885 & 0.053 \\
\textbf{all-MiniLM-L6-v2} & \cite{embeddings:all-MiniLM-L6-v2} & 0.933 & 0.323 & 0.884 & 0.050 \\
\textbf{text-embedding-3-small} & \cite{embeddings:text-embedding-3} & 1.000 & 0.362 & 0.882 & 0.051 \\
\textbf{text-embedding-3-large} & \cite{embeddings:text-embedding-3} & 0.917 & 0.463 & 0.882 & 0.054 \\
\textbf{paraphrase-MiniLM-L6-v2} & \cite{embeddings:all-MiniLM-L6-v2} & 1.150 & 0.410 & 0.870 & 0.054 \\
\textbf{bert-base-uncased} & \cite{embeddings:bert-base-uncased} & 1.600 & 0.703 & 0.777 & 0.100 \\
\textbf{spacy/en\_core\_web\_md} & \cite{embeddings:spacy} & 1.833 & 0.466 & 0.760 & 0.090 \\
\hline
\end{tabular}

\caption{Quality of different embeddings models in measuring semantic similarity between sentences. A lower average displacement and higher RBO indicate a better embeddings model for this purpose.}\label{tab:semantic}
\end{table*}
Results indicate that the all-MiniLM-L6-v2 embeddings model perform well with respect to both Average Displacement and RBO.
For this reason, we run the rest of our experiments using this embeddings model for computing the semantic similarity between sentences.

We computed the semantic similarity between all pairs of sentences in the dataset to establish a baseline for comparison.
The mean semantic similarity between a pair of unrelated sentences was $\mu \approx 0.574$ with a standard deviation of $\sigma \approx 0.061$.
Furthermore, the distribution appears to be relatively Gaussian (a histogram can be found in Appendix~\ref{apx:ss_baseline}). 
Intuitively, this suggests that semantically unrelated sentences are very unlikely to have a semantic similarity score of above $\mu + 3\sigma$ (i.e., greater than $0.757$).

A good translation, then, should score high on all three semantic similarity metrics.
For example, the translation
\displayTranslation{I am swimming.}
{I am swimming.}{1}
{I am swimming.}{1}
{Nüü pahabi-ti.}
{I am swimming.}{1}
{gpt-3.5-turbo}
\noindent is perfect.
There are other interesting cases too, though.
For example, when the comparator score is low but simple and backwards scores are high, the translator appears to do well with the vocabulary available, but is only able to give a partial translation.
For example, the translation
\displayTranslation{Birds will migrate and return.}
    {A bird will migrate. \\ The bird will return.}{0.955}
    {A bird will [VERB]. \\ The bird will [VERB].}{0.778}
    {[migrate]-wei tsiipa-uu. \\ {[return]-wei tiipa-uu}.}
    {That bird will migrate. \\ That bird will return.}{0.944}
    {gpt-4}
\noindent has a high simple score, since the ``Birds will migrate and return'' is quite similar (semantically) to ``A bird will migrate. The bird will return.''.
The comparator score is much lower, however, since the words for ``migrate'' and ``return'' are not available in the vocabulary.
The backwards score is good because the sentence ``[migrate]-wei tsiipa-uu. [return]-wei tiipa-uu.'', when translated using the OVP to English translator described in Section~\ref{sec:ovp2eng}, becomes ``That bird will migrate. That bird will return'', which is almost equivalent to the simple sentence.
This example also highlights an advantage to our approach.
No-resource language learners can use this system to understand how sentences should be structured even if it doesn't have all of the vocabulary necessary to generate the entire translation.
Furthermore, users can be fairly confident that the OVP target sentence is accurately translated into the English backwards sentence, due to the accuracy of the OVP to English translator.

The simple score being low indicates that a significant portion of the input sentence's meaning is lost during segmentation (due, for example, to the input sentence containing adverbs, adjectives, prepositions, etc. that don't fit anywhere in a simple subject-verb or subject-verb-object sentence).
For example, the translation
\displayTranslation{My brother and I went hiking.}
    {Brother went. I went.}{0.794}
    {[SUBJECT] went. I went.}{0.608}
    {mia-ku [brother]-uu. nüü mia-ku.}
    {The brother went. I went.}{0.806}
    {gpt-3.5-turbo}
\noindent lost meaning in the first step segmenting the input sentence into simple sentences because it chose to use the verb "to go" instead of "hike" which is the main topic of the sentence.
Perhaps a better way to have segmented this sentence would be: ``Brother hiked. I hiked''.
It may be possible to encourage the LLM to prefer ``topic'' verbs through prompt engineering.

Another interesting case is when the simple and comparator scores are high and only the backwards score is lower.
This is observed in cases where there is ambiguity in OVP where there is not in English.
For example, in the translation
\displayTranslation{She is cooking.}
    {She is cooking.}{1}
    {She is cooking.}{1}
    {Uhu sawa-ti.}
    {He is cooking.}{0.836}
    {gpt-4}
\noindent ``she'' turns to ``he'' in the backwards translation because OVP does not have gendered pronouns.
Despite the lower backwards score, this translation is accurate.

\begin{figure*}
    \centering
    \includegraphics[width=1\linewidth]{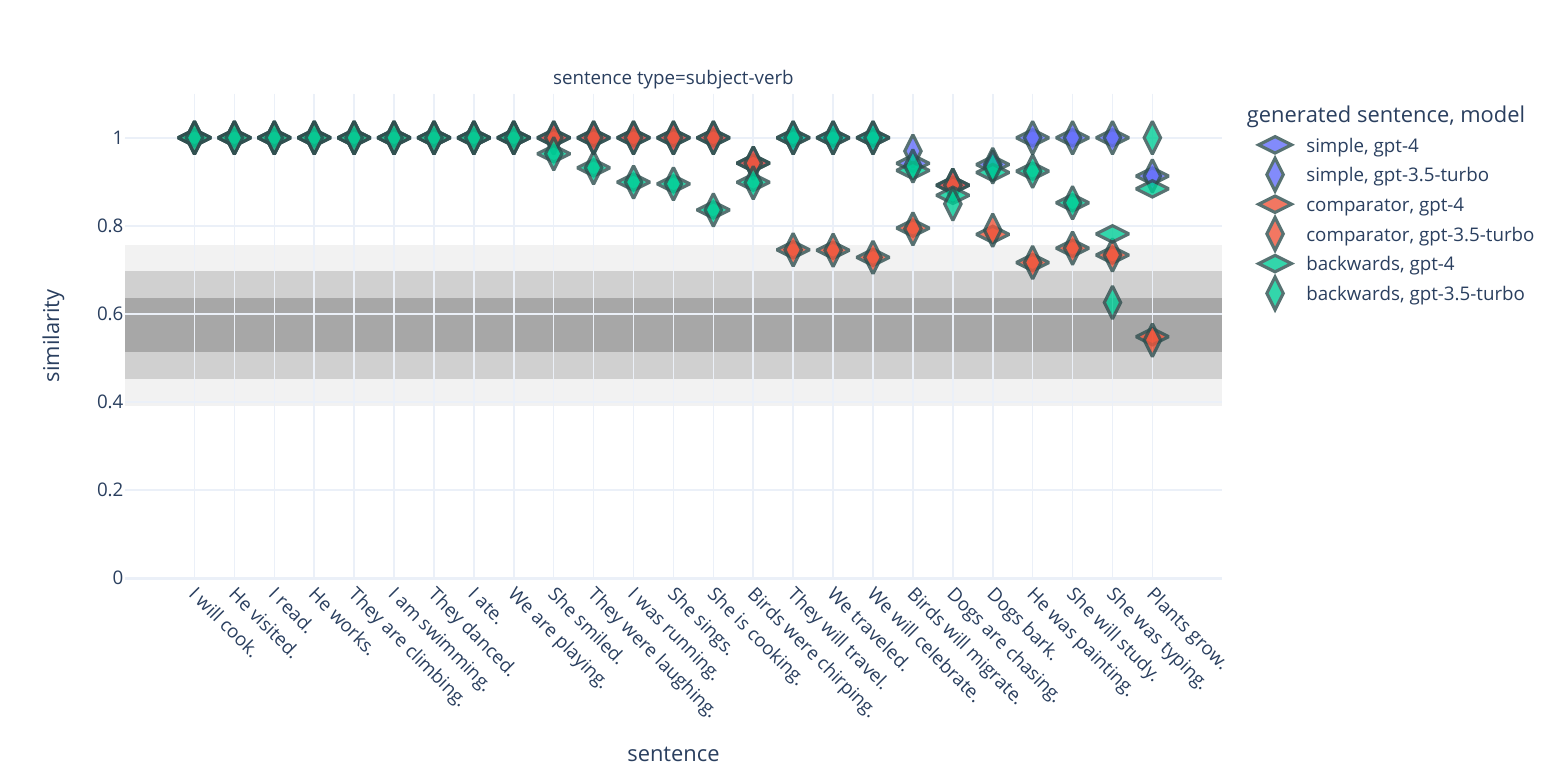}
    \caption{Results for subject-verb sentences. The dark, medium, and light gray bands represent the baseline similarity (between unrelated sentences in the dataset) +/- one, two, and three standard deviations, respectively.}
    \label{fig:result:subject-verb}
\end{figure*}

In general, both gpt-3.5-turbo and gpt-4 models do well with respect to the simple and backwards semantic similarity scores.
Table~\ref{tab:results} summarizes the mean semantic similarity scores for each model and type of sentence.
\begin{table}[ht]
    \centering
    \begin{tabular}{|llr|}
    \hline
                    &                & \textbf{Mean} \\
    \textbf{Model}  & \textbf{Type}  & \textbf{Sim.} \\
    \hline
    gpt-3.5-turbo & subject-verb                   & 0.941 \\
                  & two-verb                       & 0.906 \\
                  & subject-verb-object            & 0.869 \\
                  & two-clause                     & 0.879 \\
                  & complex                        & 0.829 \\
    gpt-4         & subject-verb                   & 0.941 \\
                  & subject-verb-object            & 0.866 \\
                  & two-verb                       & 0.905 \\
                  & two-clause                     & 0.877 \\
                  & complex                        & 0.830 \\
    \hline
    \end{tabular}
    \caption{Translation qualities by model and sentence type: mean semantic similarities between input sentence and the simple, comparator, and backwards sentences produced during translation.}
    \label{tab:results}
\end{table}
Figure~\ref{fig:result:subject-verb} depicts results for subject-verb sentences.
Plots for the rest of the results can be found in Appendix~\ref{apx:eng2ovp}.
That the simple and backwards scores are consistently higher than the comparator scores suggests that the translator can be greatly improved simply by expanding its vocabulary.
It is also interesting to note that the cheaper, weaker model performs quite well\footnote{The cost to run generate all translations was $\$0.09$ using gpt-3.5-turbo and $\$5.48$ using gpt-4.}.

\section{Conclusion}\label{sec:conclusion}
In this paper, we present the first translation tool for Owens Valley Paiute (a critically endangered Indigenous American language) and, in doing so, propose a new methodology for low/no-resource machine translation: \acronym (LLM-Assisted Rule-Based Machine Translation).
Due to a lack of bilingual or monolingual corpora, we use semantic similarity to measure the quality of translations, and observe interesting cases where the translator performs well, where it performs well given the vocabulary (primarily verbs and nouns) available to it, and where it performs poorly due to current implementation limitations.

This research opens up many directions for future work.
First, expanding the translation tool (e.g., by adding more vocabulary, building versions for other languages, etc.) will make it more useful for the communities it is meant to serve.
We are currently working to add more complex sentence structures to test the limits of this methodology.
Building this system required expert knowledge of the target language.
It may be interesting to explore how large language models might be able to help with this task as well (e.g., in finding patterns and implementing RBMT logic) to make it easier for non-experts (and even experts in the language who are less comfortable writing code) to build these kinds of translators.
Finally, other methodologies for leveraging LLMs for low/no-resource machine translation should be explored.
For example, a RAG (retrieval augmented generation)~\cite{rag} approach that searches a knowledge-base for known sentence structures, vocabulary, grammar rules, etc. and uses them to perform zero-shot translation might work well.
The remarkable general-purpose language skills that LLMs exhibit make them a promising tool in helping revitalize critically endangered languages.

\FloatBarrier
\bibliography{main}

\newpage
\appendix
\clearpage
\onecolumn
\appendixpage
\section{OVP to English Translator Results}\label{apx:ovp2eng}

\begin{longtable}{|c|c|c|}
\hline
\textbf{sentence} & \textbf{translation} & \textbf{label} \\ \hline
tei-zawa-ku ihi & This cooked us. & 1 \\ \hline
isha'-uu tüba-neika ai-hibi-pü & That coyote has drunk these pinenuts. & 1 \\ \hline
tüba-uu tsibui-dü & Those pinenuts climb. & 1 \\ \hline
ta-\~{w}ui-gaa-wei mahu\~{w}a & "They are going to write to us, you and I." & 1 \\ \hline
wo'ada-uu aaponu'-oka u-zawa-ti & The mosquito is cooking the apple. & 1 \\ \hline
tübbi-uu tüwoobü-neika a-yadohi-pü & The rock has talked to the earth. & 1 \\ \hline
payahuupü-uu katü-ti & That river is sitting. & 1 \\ \hline
kwisha'i-wei üü & You will sneeze. & 1 \\ \hline
isha'-ii tübbi-neika mai-\~{w}ui-gaa-wei & This coyote is going to write these rocks. & 1 \\ \hline
toni-uu wünü-ti & The wickiup is standing. & 1 \\ \hline
isha'pugu-neika ihi mai-dama'i-ku & This found these dogs. & 1 \\ \hline
wo'ada-neika ihi mai-dama'i-gaa-wei & This will find these mosquitoes. & 1 \\ \hline
tabuutsi'-uu tüba-noka u-buni-ku & The cottontail saw those pinenuts. & 1 \\ \hline
maishibü-neika uhu ai-nia-ti & He/she/it is reading these corn. & 1 \\ \hline
koopi'-ii wükada-noka ui-nia-ku & This coffee read those bird snakes. & 1 \\ \hline
tuunapi-uu waakü-pü & That food has worked. & 1 \\ \hline
katünu-ii koopi'-oka ui-nobini-wei & This chair will visit those coffees. & 1 \\ \hline
aingwü-neika mahu ma-nia-dü & He/she/it reads the squirrel. & 1 \\ \hline
maishibü-uu wükihaa-gaa-wei & That corn is going to smile. & 1 \\ \hline
isha'-uu isha'-oka ui-zawa-wei & That coyote will cook those coyotes. & 1 \\ \hline
aingwü-ii tsibui-dü & This squirrel climbs. & 1 \\ \hline
katünu-ii tübbi-neika ma-buni-wei & This chair will see this rock. & 1 \\ \hline
wükada-uu tsibui-ku & The bird snake climbed. & 1 \\ \hline
wo'ada-uu paya-neika ma-hibi-ti & The mosquito is drinking the water. & 1 \\ \hline
pagwi-neika mahu ma-\~{w}ui-pü & He/she/it has written/is writing this fish. & 1 \\ \hline
tabuutsi'-uu isha'pugu-neika mai-nobini-gaa-wei & That cottontail is going to visit those dogs. & 1 \\ \hline
paya-neika mahu mai-hibi-gaa-wei & He/she/it is going to drink this water. & 1 \\ \hline
nishua'i-pü nüü & I am laughing. & 1 \\ \hline
aaponu'-ii küna-neika a-düka-pü & This apple has eaten this wood. & 1 \\ \hline
katü-dü uhu & He/she/it sits. & 1 \\ \hline
mukita-uu isha'pugu-noka u-naki-ti & The lizard is chasing the dog. & 1 \\ \hline
isha'-oka üü ui-dama'i-ku & You found those coyotes. & 1 \\ \hline
pahabichi-uu wükihaa-dü & That bear smiles. & 1 \\ \hline
pahabichi-ii wo'abi-noka ui-naka-dü & This bear hears those worms. & 1 \\ \hline
habi-ku ihi & This lay down. & 1 \\ \hline
tümui-ku taa & You and I wrote. & 1 \\ \hline
üwi-ku ihi\~{w}a & These slept. & 1 \\ \hline
tübbi-uu ta-naka-ku & "That rock heard us, you and I." & 1 \\ \hline
wo'ada-uu tei-gwana-dü & That mosquito smells us. & 1 \\ \hline
tümui-dü mahu\~{w}a & They write. & 1 \\ \hline
üwi-dü nüügwa & We are sleeping. & 1 \\ \hline
tsibui-ti mahu & He/she/it is climbing. & 1 \\ \hline
mukita-uu tsibui-pü & That lizard has climbed. & 1 \\ \hline
payahuupü-uu toyabi-neika ma-zawa-gaa-wei & The river is going to cook the mountain. & 1 \\ \hline
nobi-uu kwisha'i-ku & That house sneezed. & 1 \\ \hline
kamü-uu wükihaa-dü & That jackrabbit smiles. & 1 \\ \hline
toni-uu katü-wei & That wickiup will sit. & 1 \\ \hline
aingwü-uu katünu-noka u-zawa-gaa-wei & The squirrel is going to cook that chair. & 1 \\ \hline
paya-uu pasohobü-neika ai-buni-wei & That water will see those trees. & 1 \\ \hline
toyabi-ii tsibui-ku & The mountain climbed. & 1 \\ \hline
tsibui-wei taa & You and I will climb. & 1 \\ \hline
pugu-uu wo'abi-neika ai-naki-ku & That horse chased those worms. & 1 \\ \hline
mukita-uu wai-noka u-nobini-gaa-wei & The lizard is going to visit the rice. & 1 \\ \hline
wükihaa-ti mahu & He/she/it is smiling. & 1 \\ \hline
tüsüga-ii tüwoobü-neika ma-naki-gaa-wei & This weasel is going to chase this earth. & 1 \\ \hline
yadoha-ku uhu\~{w}a & They talked. & 1 \\ \hline
pahabichi-ii pugu-noka ui-nia-ku & The bear read those horses. & 1 \\ \hline
paya-uu katünu-noka ui-yadohi-gaa-wei & Water is going to talk to those chairs. & 1 \\ \hline
pagwi-ii wo'abi-noka ui-düka-ti & This fish is eating those worms. & 1 \\ \hline
tabuutsi'-uu tübinohi-ku & That cottontail played. & 1 \\ \hline
tünia-ku nüü & I read. & 1 \\ \hline
poyoha-pü ihi\~{w}a & These are running. & 0 \\ \hline
mukita-uu yotsi-wei & That lizard will fly. & 1 \\ \hline
tabuutsi'-uu tübbi-neika ma-buni-pü & That cottontail has seen this rock. & 1 \\ \hline
isha'pugu-ii tüba-neika ai-nobini-ku & This dog visited these pinenuts. & 1 \\ \hline
isha'-uu katü-dü & That coyote sits. & 1 \\ \hline
pasohobü-ii tsiipa-noka ui-naka-ku & The tree heard those birds. & 1 \\ \hline
kamü-uu mukita-neika ma-zawa-wei & That jackrabbit will cook this lizard. & 1 \\ \hline
tuunapi-uu tümui-gaa-wei & That food is going to write. & 1 \\ \hline
wai-uu aingwü-neika ai-naka-pü & That rice has heard those squirrels. & 1 \\ \hline
tsiipa-uu pugu-noka ui-naka-pü & The bird has heard those horses. & 1 \\ \hline
pagwi-ii wükihaa-gaa-wei & These fish are going to smile. & 1 \\ \hline
tei-nobini-dü uhu\~{w}a & They visit us. & 1 \\ \hline
paya-neika ihi ma-dama'i-dü & This finds this water. & 1 \\ \hline
isha'pugu-neika nüügwa ma-düka-ku & We ate this dog. & 1 \\ \hline
tabuutsi'-uu hubiadu-dü & The cottontail sings. & 1 \\ \hline
kwadzi-ii yadoha-wei & This tail will talk. & 1 \\ \hline
isha'pugu-uu mukita-neika ai-naka-ti & That dog is hearing these lizards. & 1 \\ \hline
pasohobü-noka üü u-zawa-gaa-wei & You are going to cook that tree. & 1 \\ \hline
tsiipa-noka uhu u-buni-pü & He/she/it has seen or is seeing that bird. & 1 \\ \hline
tuunapi-neika mahu mai-gwati-ku & He/she/it hit the foods here. & 0 \\ \hline
isha'-eika nüügwa ai-\~{w}ui-gaa-wei & We are going to write coyotes. & 1 \\ \hline
pasohobü-uu toyabi-neika ma-hibi-pü & The tree has drunk the mountain. & 1 \\ \hline
aingwü-uu aaponu'-eika ai-naka-ti & The squirrel is hearing these apples. & 1 \\ \hline
tabuutsi'-uu wai-noka u-yadohi-pü & The cottontail has talked to the rice. & 1 \\ \hline
katünu-noka mahu\~{w}a ui-gwati-dü & They are hitting those chairs. & 1 \\ \hline
tüwoobü-neika uhu\~{w}a mai-buni-ti & They are seeing these earths. & 1 \\ \hline
koopi'-uu huka\~{w}ia-ti & Coffee is walking. & 1 \\ \hline
küna-ii ni-naka-ti & This wood is hearing us. & 1 \\ \hline
toyabi-neika taagwa ma-buni-gaa-wei & We are going to see this mountain. & 1 \\ \hline
isha'pugu-ii üwi-dü & This dog sleeps. & 1 \\ \hline
pagwi-neika ihi mai-naka-gaa-wei & This will hear these fish. & 1 \\ \hline
kwadzi-ii toni-neika ai-gwati-gaa-wei & This tail is going to hit those wickiups. & 1 \\ \hline
paya-ii tuunapi-noka u-düka-ti & This water is eating that food. & 1 \\ \hline
mukita-uu tümui-gaa-wei & That lizard is going to write. & 1 \\ \hline
pahabichi-uu küna-neika ma-zawa-ku & The bear cooked the wood. & 1 \\ \hline
isha'pugu-uu tabuutsi'-eika a-zawa-dü & That dog is cooking this cottontail. & 1 \\ \hline
katünu-uu pahabichi-noka u-naki-ku & The chair chased the bear. & 1 \\ \hline
küna-uu waakü-gaa-wei & That wood is going to work. & 1 \\ \hline
pugu-neika mahu ai-naka-ku & He/she/it heard these horses. & 1 \\ \hline

\caption{One hundred randomly generated sentences (using the OVP sentence building tool described in Section~\ref{sec:ovp2eng}) and translated (using the OVP to English translator described in Section~\ref{sec:ovp2eng}) labeled 1 if the translation is accurate and 0 otherwise.}
\label{tab:ovp2eng:accuracy}
\end{longtable}

\FloatBarrier
\section{Semantic Similarity Sentences}\label{apx:semantic_similarity}
\begin{longtable}{|l|p{0.6\textwidth}|}
\hline
\textbf{Base Sentence} & \textbf{Other Sentences \newline {\footnotesize (in order of most to least semantically similar to Base Sentence)}} \\
\hline
She sings. & He sings. \\
& He/she/it sings. \\
& She performs a song. \\
& A song is being sung by her. \\
& She hums a tune. \\
& She listens to music. \\
& She dances. \\
& She eats. \\
& The cat sleeps. \\
& Mountains echo silently. \\
\hline
The dog fell. & The dog fell yesterday. \\
& A dog stumbled. \\
& The puppy tripped over. \\
& The cat is running. \\
& An animal is in motion. \\
& The bird flies. \\
& Leaves fall in autumn. \\
& He reads a book. \\
& Clouds cover the sky. \\
& Apples on the moon are hungry. \\
\hline
The man ate an apple. & The apple was eaten by the man. \\
& A man consumes a fruit. \\
& The boy nibbles on an apple. \\
& Someone is eating. \\
& He drinks water. \\
& The woman ate a pie. \\
& A cat chases a mouse. \\
& Trees grow in the forest. \\
& The car is red. \\
& Stars twinkle at night. \\
\hline
The sun rises in the east. & The east welcomes the sunrise. \\
& Sunrise occurs in the east. \\
& Day breaks in the east. \\
& The moon sets in the west. \\
& The stars shine at night. \\
& Clouds gather before rain. \\
& The wind changes direction. \\
& Leaves fall in autumn. \\
& Snow covers the mountains. \\
& A book rests on the table. \\
\hline
Birds fly south for the winter. & For winter, birds head south. \\
& Migratory birds travel south when it gets cold. \\
& Birds migrate to warmer climates during winter. \\
& Fish swim upstream. \\
& Bears hibernate in winter. \\
& Flowers bloom in spring. \\
& The earth orbits the sun. \\
& Trees lose their leaves in fall. \\
& The sky is blue. \\
& A cat sleeps on the couch. \\
\hline
I read a book yesterday. & Yesterday, I finished reading a book. \\
& A book was read by me yesterday. \\
& I watched a movie last night. \\
& I'll visit the library tomorrow. \\
& She writes a letter. \\
& He cooks dinner. \\
& They are painting a house. \\
& The sun sets in the evening. \\
& A dog barks at night. \\
& The car needs fuel. \\
\hline
The cake was delicious. & Delicious was the cake. \\
& The dessert tasted great. \\
& We enjoyed the tasty cake. \\
& The pie is sour. \\
& Coffee complements breakfast. \\
& Leaves rustle in the wind. \\
& A bird sings outside. \\
& Children play in the park. \\
& Traffic is heavy today. \\
& The phone is ringing. \\
\hline
Lightning precedes thunder. & Thunder follows lightning. \\
& First comes lightning, then comes thunder. \\
& The storm brings lightning and thunder. \\
& Rain refreshes the earth. \\
& The sun warms the ground. \\
& A river flows to the sea. \\
& Mountains reach towards the sky. \\
& A cat chases a mouse. \\
& Books fill the shelf. \\
& The clock ticks steadily. \\
\hline
She painted a beautiful picture. & A beautiful picture was painted by her. \\
& The painting she created is beautiful. \\
& She sketches a portrait. \\
& He writes a poem. \\
& They are filming a movie. \\
& Birds nest in spring. \\
& Flowers wilt in the heat. \\
& Kids play video games. \\
& Cars fill the parking lot. \\
& The sun sets late in summer. \\
\hline
The computer is broken. & A broken state afflicts the computer. \\
& The machine isn't working. \\
& We need to repair the computer. \\
& The phone's battery is dead. \\
& Lights flicker during a power outage. \\
& A book lies open on the desk. \\
& Water boils at 100 degrees Celsius. \\
& A cat purrs contentedly. \\
& The door creaks when opened. \\
& Birds migrate in autumn. \\
\hline
He solved the puzzle quickly. & The puzzle was quickly solved by him. \\
& Quickly, he found the solution to the puzzle. \\
& She completes the crossword. \\
& The mystery remains unsolved. \\
& A race against time. \\
& Flowers are sold at the market. \\
& The river cuts through the valley. \\
& A key unlocks the door. \\
& Leaves turn red in autumn. \\
& The train arrives at noon. \\
\hline
The stars twinkle at night. & At night, the stars shimmer. \\
& Twinkling stars fill the night sky. \\
& Night unveils a sky full of stars. \\
& The moon glows brightly. \\
& Clouds mask the moon. \\
& The sun sets, stars appear. \\
& A comet streaks through the sky. \\
& Fireflies glow in the dark. \\
& Crickets chirp in the evening. \\
& A candle flickers in the window. \\
\hline
\caption{Base sentences and other sentences ordered by their semantic similarity to the base sentence (as determined by authors).}
\label{tab:apx:semantic_similarity}
\end{longtable}

\FloatBarrier
\section{Vocabulary}\label{apx:vocab}
\begin{table}[htbp]
    \begin{subtable}{0.33\textwidth}
        \centering
        \begin{tabular}{r|l}
            \textit{tüka} & eat \\
            \textit{puni} & see \\
            \textit{hibi} & drink \\
            \textit{naka} & hear \\
            \textit{kwana} & smell \\
            \textit{kwati} & hit \\
            \textit{yadohi} & talk to \\
            \textit{naki} & chase \\
            \textit{tsibui} & climb \\
            \textit{sawa} & cook \\
            \textit{tama'i} & find \\
            \textit{nia} & read \\
            \textit{mui} & write \\
            \textit{nobini} & visit \\
        \end{tabular}
        \caption{Transitive Verbs}
        \label{tab:transitive_verbs}
        \vspace{0.5cm} 
        \begin{tabular}{r|l}
            \textit{katü} & sit \\
            \textit{üwi} & sleep \\
            \textit{kwisha'i} & sneeze \\
            \textit{poyoha} & run \\
            \textit{mia} & go \\
            \textit{huka\~{w}a} & walk \\
            \textit{wünü} & stand \\
            \textit{habi} & lie down \\
            \textit{yadoha} & talk \\
            \textit{kwatsa'i} & fall \\
            \textit{waakü} & work \\
            \textit{wükihaa} & smile \\
            \textit{hubiadu} & sing \\
            \textit{nishua'i} & laugh \\
            \textit{tsibui} & climb \\
            \textit{tübinohi} & play \\
            \textit{yotsi} & fly \\
            \textit{nüga} & dance \\
            \textit{pahabi} & swim \\
            \textit{tünia} & read \\
            \textit{tümui} & write \\
            \textit{tsiipe'i} & chirp \\
        \end{tabular}
        \caption{Intransitive Verbs}
        \label{tab:intransitive_verbs}
    \end{subtable}%
    \begin{subtable}{0.33\textwidth}
        \centering
        \begin{tabular}{r|l}
            \textit{isha'} & coyote \\
            \textit{isha'pugu} & dog \\
            \textit{kidi'} & cat \\
            \textit{pugu} & horse \\
            \textit{wai} & rice \\
            \textit{tüba} & pinenuts \\
            \textit{maishibü} & corn \\
            \textit{paya} & water \\
            \textit{payahuupü} & river \\
            \textit{katünu} & chair \\
            \textit{toyabi} & mountain \\
            \textit{tuunapi} & food \\
            \textit{pasohobü} & tree \\
            \textit{nobi} & house \\
            \textit{toni} & wickiup \\
            \textit{apo} & cup \\
            \textit{küna} & wood \\
            \textit{tübbi} & rock \\
            \textit{tabuutsi'} & cottontail \\
            \textit{kamü} & jackrabbit \\
            \textit{aaponu'} & apple \\
            \textit{tüsüga} & weasle \\
            \textit{mukita} & lizard \\
            \textit{wo'ada} & mosquito \\
            \textit{wükada} & bird snake \\
            \textit{wo'abi} & worm \\
            \textit{aingwü} & squirrel \\
            \textit{tsiipa} & bird \\
            \textit{tüwoobü} & earth \\
            \textit{koopi'} & coffee \\
            \textit{pahabichi} & bear \\
            \textit{pagwi} & fish \\
            \textit{kwadzi} & tail \\
        \end{tabular}
        \caption{Nouns}
        \label{tab:nouns}
        \vspace{0.5cm} 
        \begin{tabular}{r|l}
            \textit{ku} & completive (past) \\
            \textit{ti} & present ongoing (-ing) \\
            \textit{dü} & present \\
            \textit{wei} & future (will) \\
            \textit{gaa-wei} & future (going to) \\
            \textit{pü} & have x-ed, am x-ed \\
        \end{tabular}
        \caption{Object Suffixes}
        \label{tab:verb_tenses}
    \end{subtable}%
    \begin{subtable}{0.33\textwidth}
        \centering
        \begin{tabular}{r|l}
            \textit{nüü} & I \\
            \textit{uhu} & he/she/it \\
            \textit{uhu\~{w}a} & they \\
            \textit{mahu} & he/she/it \\
            \textit{mahu\~{w}a} & they \\
            \textit{ihi} & this \\
            \textit{ihi\~{w}a} & these \\
            \textit{taa} & you and I \\
            \textit{nüügwa} & we (exclusive) \\
            \textit{taagwa} & we (inclusive) \\
            \textit{üü} & you \\
            \textit{üügwa} & you (plural) \\  
        \end{tabular}
        \caption{Subject Pronouns}
        \label{tab:subject_pronouns}
        \vspace{0.5cm} 
        \begin{tabular}{r|l}
            \textit{ii} & (proximal) \\
            \textit{uu} & (distal) \\
        \end{tabular}
        \caption{Subject Suffixes}
        \label{tab:subject_suffixes}
        \vspace{0.5cm} 
        \begin{tabular}{r|l}
            \textit{i} & me \\
            \textit{u} & him/her/it (distal) \\
            \textit{ui} & them (distal) \\
            \textit{ma} & him/her/it (proximal) \\
            \textit{mai} & them (proximal) \\
            \textit{a} & him/her/it (proximal) \\
            \textit{ai} & them (proximal) \\
            \textit{ni} & us (plural, exclusive) \\
            \textit{tei} & us (plural, inclusive) \\
            \textit{ta} & us (dual), you and I \\
            \textit{ü} & you (singular) \\
            \textit{üi} & you (plural), you all \\
        \end{tabular}
        \caption{Object Pronouns}
        \label{tab:object_pronouns}
        \vspace{0.5cm} 
        \begin{tabular}{r|l}
            \textit{eika} & (proximal) \\
            \textit{oka} & (distal) \\
        \end{tabular}
        \caption{Object Suffixes}
        \label{tab:object_suffixes}
    \end{subtable}
    \caption{Vocabulary available in sentence building system.}
    \label{tab:vocabulary}
\end{table}

\FloatBarrier
\section{English to OVP Translation Results}\label{apx:eng2ovp}

\begin{figure*}
    \centering
    \includegraphics[width=1\linewidth]{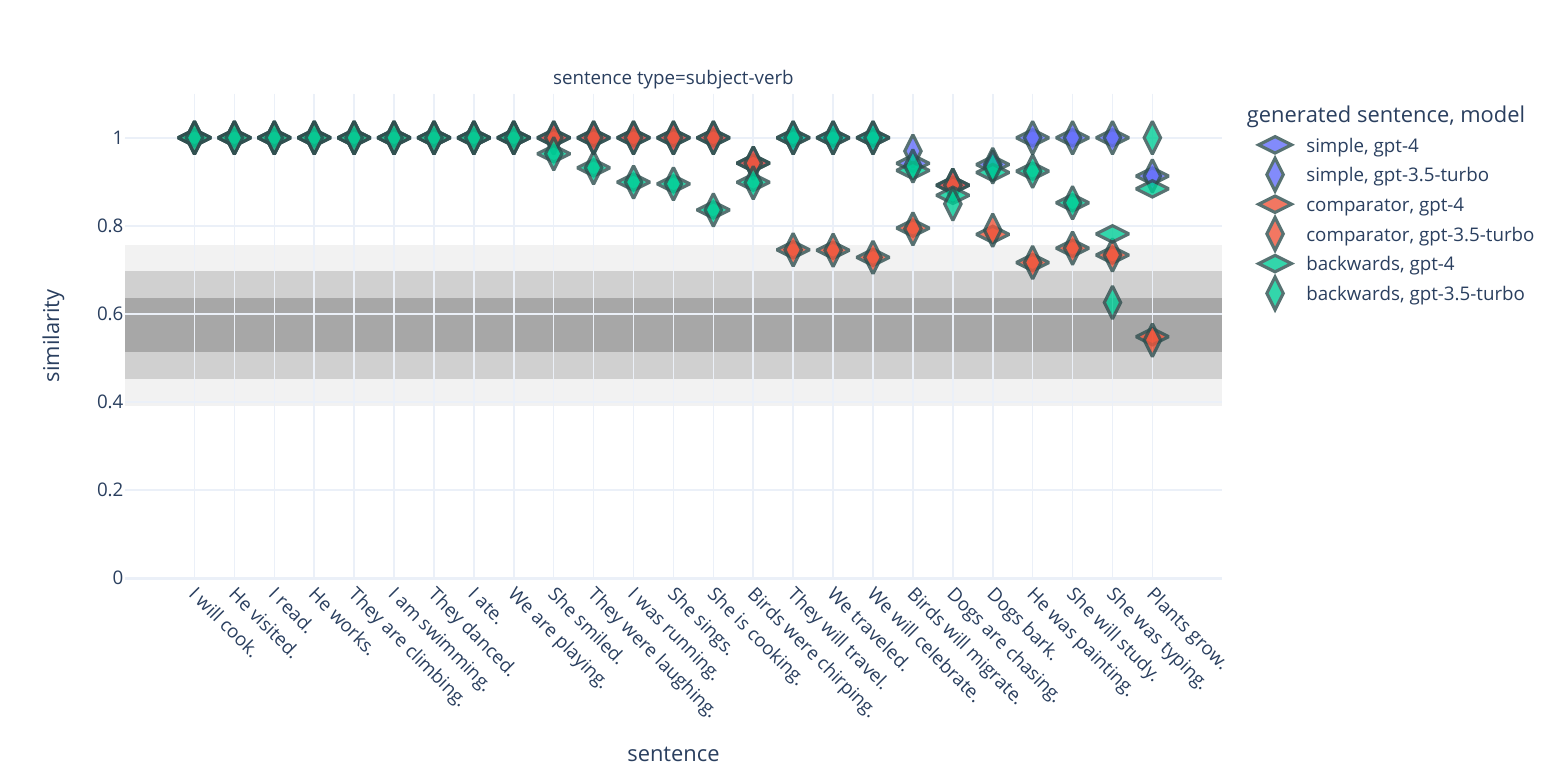}
    \caption{Results for subject-verb sentences. The dark, medium, and light gray bands represent the baseline similarity (between unrelated sentences in the dataset) +/- one, two, and three standard deviations, respectively.}
    \label{fig:apx:result:subject-verb}
\end{figure*}

\begin{figure*}
    \centering
    \includegraphics[width=1\linewidth]{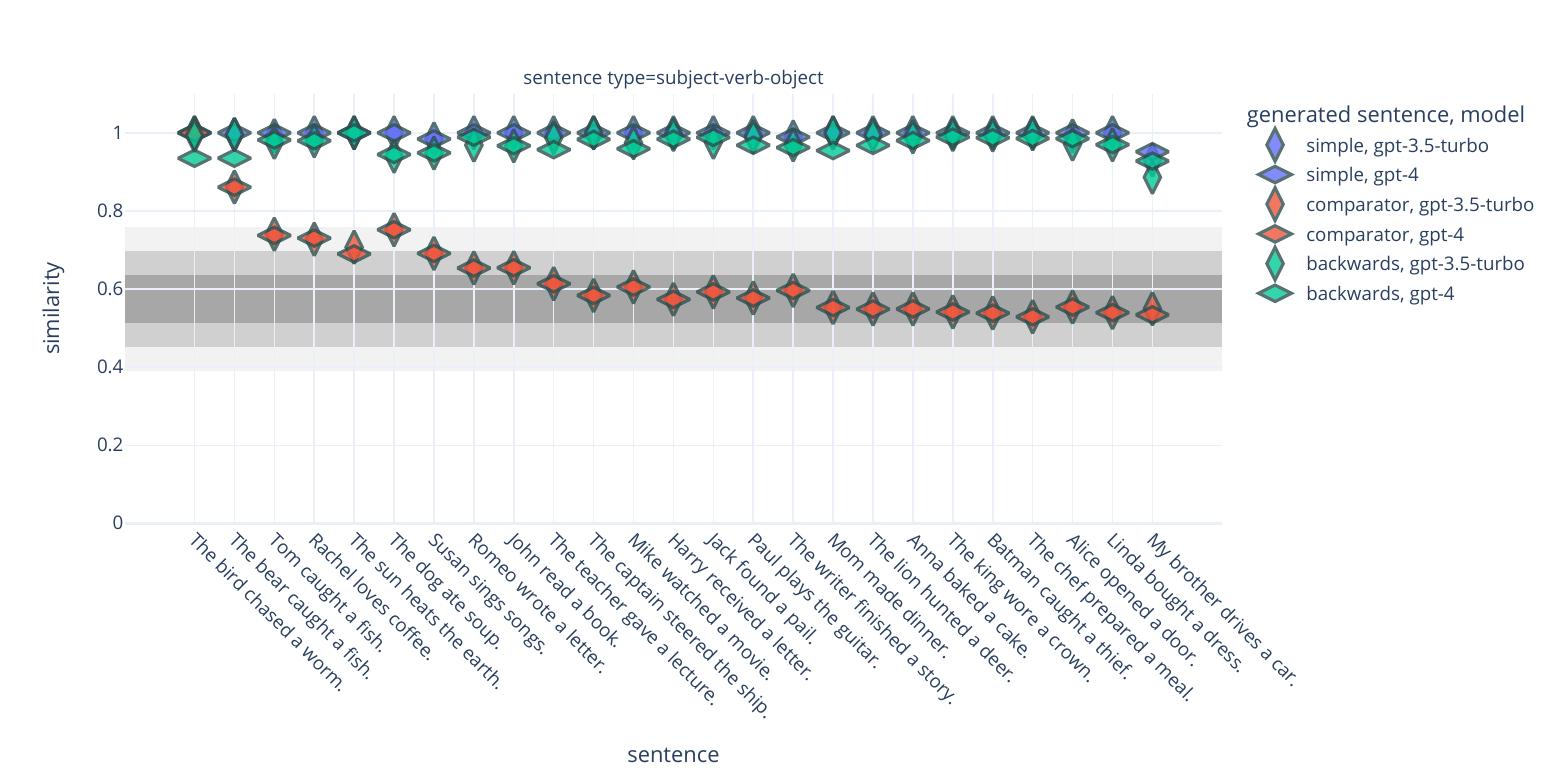}
    \caption{Results for subject-verb-object sentences. The dark, medium, and light gray bands represent the baseline similarity (between unrelated sentences in the dataset) +/- one, two, and three standard deviations, respectively.}
    \label{fig:apx:result:subject-verb-object}
\end{figure*}

\begin{figure*}
    \centering
    \includegraphics[width=1\linewidth]{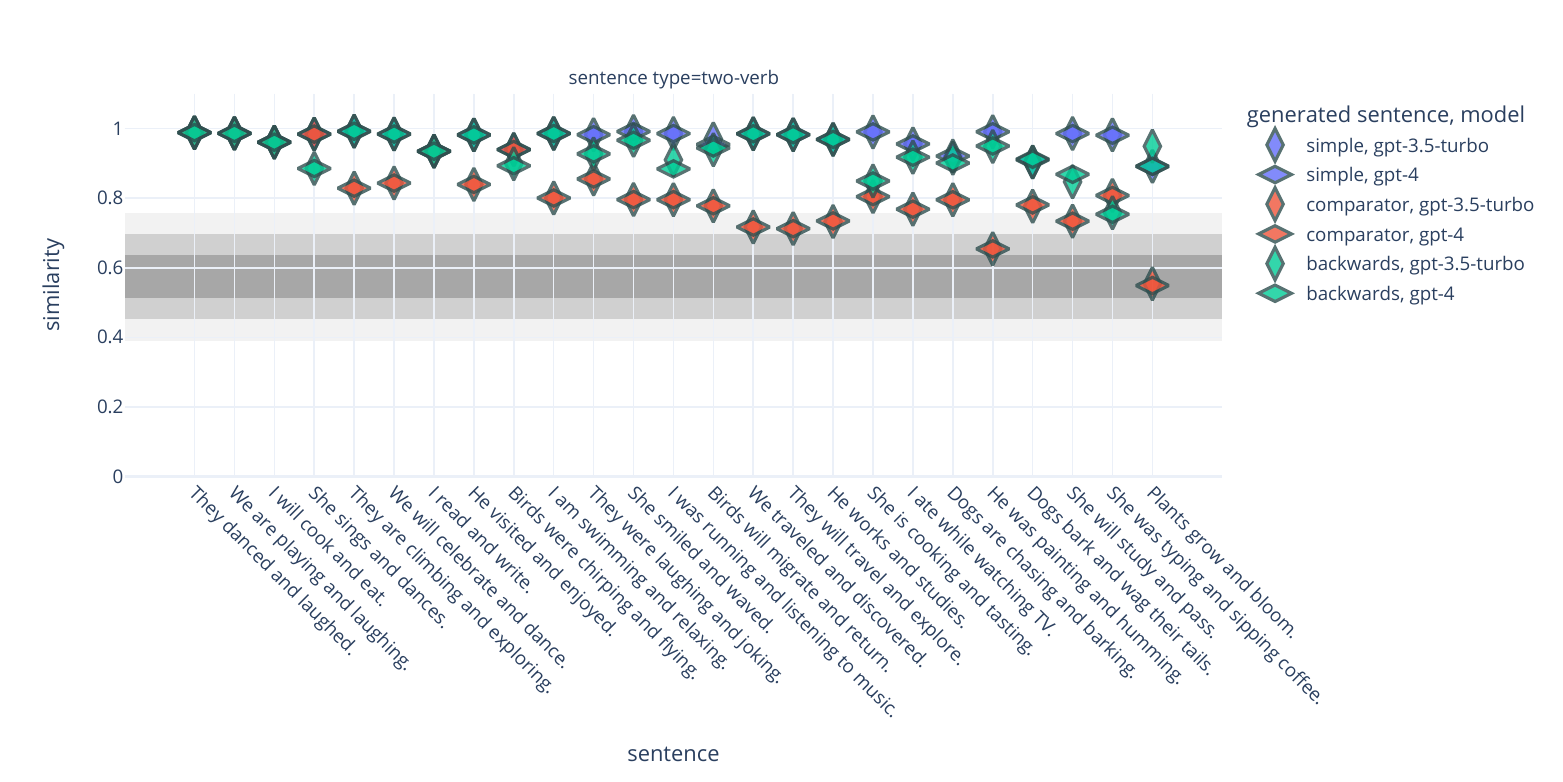}
    \caption{Results for two-verb sentences. The dark, medium, and light gray bands represent the baseline similarity (between unrelated sentences in the dataset) +/- one, two, and three standard deviations, respectively.}
    \label{fig:apx:result:two-verb}
\end{figure*}

\begin{figure*}
    \centering
    \includegraphics[width=1\linewidth]{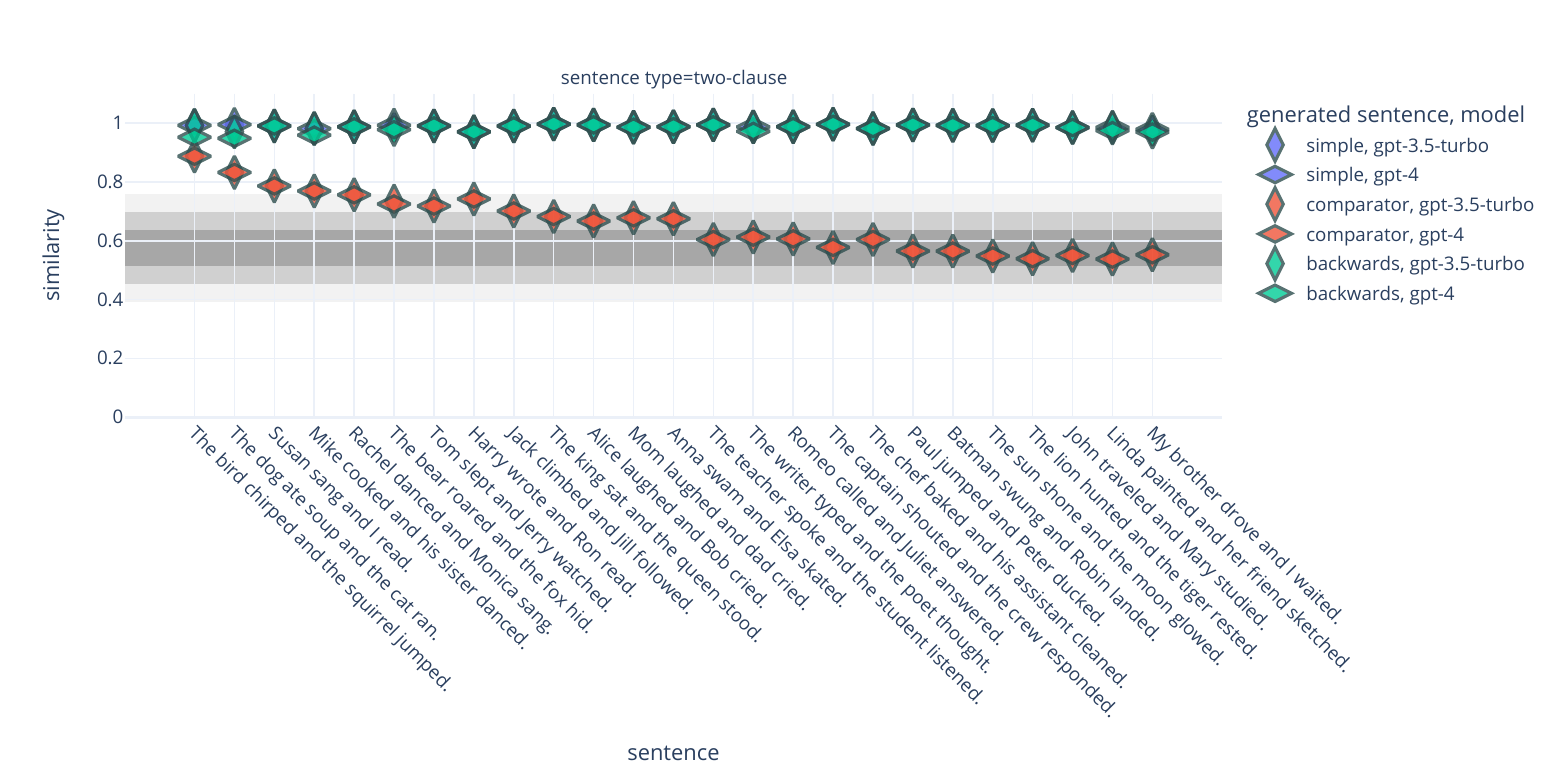}
    \caption{Results for two-clause sentences. The dark, medium, and light gray bands represent the baseline similarity (between unrelated sentences in the dataset) +/- one, two, and three standard deviations, respectively.}
    \label{fig:apx:result:two-clause}
\end{figure*}

\begin{figure*}
    \centering
    \includegraphics[width=1\linewidth]{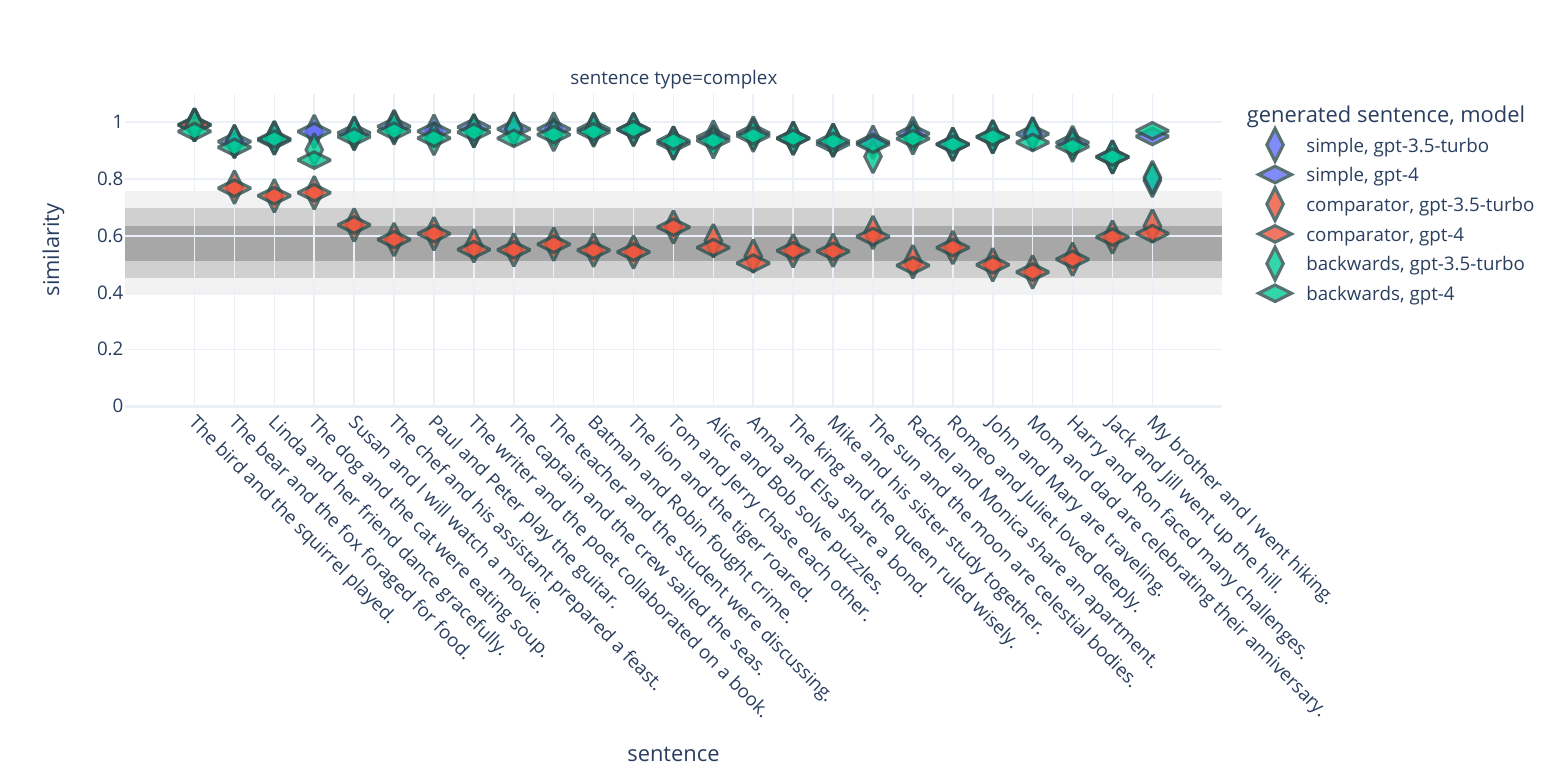}
    \caption{Results for complex sentences. The dark, medium, and light gray bands represent the baseline similarity (between unrelated sentences in the dataset) +/- one, two, and three standard deviations, respectively.}
    \label{fig:apx:result:complex}
\end{figure*}

\FloatBarrier
\section{Semantic Similarity Baseline Score}\label{apx:ss_baseline}
\begin{figure}
    \centering
    \includegraphics[width=\linewidth]{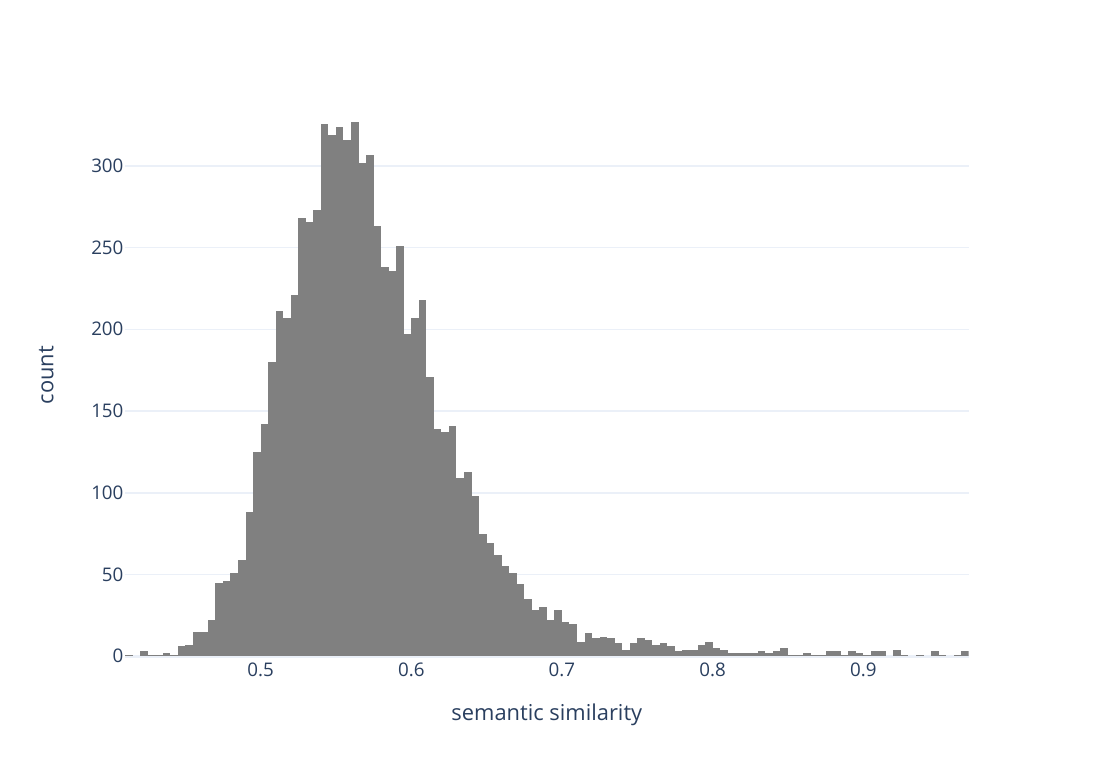}
    \caption{Distribution of semantic similarity scores between all pairs of sentences in the dataset.}
    \label{fig:apx:baseline_similarities}
\end{figure}

\FloatBarrier

\clearpage
\twocolumn

\end{document}